\documentclass[conference]{IEEEtran}
\IEEEoverridecommandlockouts

\usepackage{graphicx}
\usepackage{subcaption}
\usepackage{hyperref}

\usepackage{cite}
\usepackage{url} 
\usepackage{amsmath,amssymb,amsfonts}
\usepackage{algorithmic}
\usepackage{graphicx}
\usepackage{textcomp}
\usepackage{xcolor}
\def\BibTeX{{\rm B\kern-.05em{\sc i\kern-.025em b}\kern-.08em
    T\kern-.1667em\lower.7ex\hbox{E}\kern-.125emX}}
\begin{document}

\title{Integrating Pre-trained Language Model into Neural Machine Translation}

\author{\IEEEauthorblockN{Soon-Jae Hwang}
\IEEEauthorblockA{\textit{Korea University} \\
Seoul, Republic of Korea \\
vhch77@korea.ac.kr}
\and
\IEEEauthorblockN{Chang-Sung Jeong}
\IEEEauthorblockA{\textit{Korea University} \\
Seoul, Republic of Korea \\
csjeong@korea.ac.kr}}

\maketitle

\begin{abstract}
Neural Machine Translation (NMT) has become a significant technology in natural language processing through extensive research and development. However, the deficiency of high-quality bilingual language pair data still poses a major challenge to improving NMT performance. Recent studies have been exploring the use of contextual information from pre-trained language model (PLM) to address this problem. Yet, the issue of incompatibility between PLM and NMT model remains unresolved. This study proposes PLM-integrated NMT (PiNMT) model to overcome the identified problems. PiNMT model consists of three critical components, PLM Multi Layer Converter, Embedding Fusion, and Cosine Alignment, each playing a vital role in providing effective PLM information to NMT. Furthermore, two training strategies, Separate Learning Rates and Dual Step Training, are also introduced in this paper. By implementing the proposed PiNMT model and training strategy, we achieve state-of-the-art performance on the IWSLT'14 En$\leftrightarrow$De dataset. This study's outcomes are noteworthy as they demonstrate a novel approach for efficiently integrating PLM with NMT to overcome incompatibility and enhance performance.
\end{abstract}


\section{Introduction}
 Neural Machine Translation (NMT) has emerged as a prominent research topic in artificial intelligence and natural language processing over recent years. Particularly, Transformer model \cite{vaswani2017attention} utilizing the Attention mechanism has played a decisive role in substantially enhancing the performance of NMT. However, there remain several challenges in training NMT model. One primary challenge is the requirement of vast amounts of high-quality bilingual pair language data. Collecting and curating such data entail significant costs and time. Following previous studies \cite{edunov2019pretrained, koehn2017six}, the absence of high-quality bilingual pair language data complicates the training of NMT model and leads to performance deterioration.

Against this backdrop, Pre-trained Language Models (PLMs) such as ELMo \cite{peters2018elmo}, GPT \cite{radford2018gpt}, BERT \cite{devlin2019bert}, XLNet \cite{yang2019xlnet}, BART \cite{lewis2019bart}, and T5 \cite{raffel2020texttotext} acquire rich contextual information from readily available large-scale monolingual data. Leveraging this information, they undergo fine-tuning for downstream tasks and have achieved impressive results on key natural language processing benchmarks like GLUE \cite{wang2018glue} and SUPERGLUE \cite{sarlin2020superglue}.

Given the evident superior effects of fine-tuning PLMs for downstream tasks, we have delved into existing research to understand the potential integration of PLM into NMT model. Some explored methods include: initializing model parameters with PLM checkpoint instead of random initialization followed by fine-tuning \cite{Ramachandran2016, Lample2019, clinchant2019bertnmt, rothe2020leveraging, Ma2020}; indirectly employing PLM output in NMT through distillation \cite{Yang2019, weng2020acquiring}; and directly utilizing PLM output as input for NMT model \cite{clinchant2019bertnmt, zhu2020incorporating, weng2020acquiring, xu2021bibert}.

However, incorporating PLM into NMT is not straightforward. Fine-tuning the PLM resulted in relatively lower performance compared to using the output of a frozen PLM as input for NMT \cite{zhu2020incorporating}. The reason is the occurrence of Catastrophic Forgetting \cite{goodfellow2013empirical} during the process of transferring pre-existing knowledge from PLM to NMT. On the other hand, not fine-tuning also led to decreased performance \cite{clinchant2019bertnmt}. This drop is due to the incompatibility arising from differences in the training task, model structure, and domain of train data between PLM and NMT. For instance, while PLMs like BERT \cite{devlin2019bert} operate with an encoder structure restoring masked monolingual language data, NMT models employ an encoder-decoder structure, translating source language data to target language data. In conclusion, a new strategy is needed to overcome the issues identified above.

This paper presents a novel PLM-integrated NMT (PiNMT) Model as a solution to the previously identified challenges, effectively merging PLM and NMT. The PiNMT model is composed of three primary components: PLM Multi Layer Converter that effectively transforms the deep and rich multi-layer contextual information from PLM into information suitable for NMT; Embedding Fusion that addresses the complex fine-tuning issues of PLM; and Cosine Alignment, which prevents potential information loss during the information transfer process between the two models. 
Additionally, to enhance the efficiency and accuracy of model training, we introduce two strategies: Separate Learning Rates, applying different learning rates considering the complexity and scale between PLM and NMT model; Dual Step Training, which further amplifies model performance through the use of bidirectional data. The code implementation is publicly accessible at the following repository\footnote{Available at: \url{https://github.com/vhch/PiNMT}}.

Through these strategically designed approaches, our model exhibits a remarkable improvement on the IWSLT'14 En$\leftrightarrow$De dataset, showcasing a 5.16 BLEU score increase compared to the basic model. Notably, this result surpasses the previously highest-performing model on the same dataset by an additional 1.55 BLEU score, thereby solidifying its superior performance.

\section{Related Work}
\subsection{Pretrained Language Model}
In the field of NLP, various PLMs have been proposed to exploit large-scale monolingual data across different languages and domains. Mikolov et al. \cite{Mikolov2013} introduced two architectures, CBOW and Skip-Gram, effectively learning word vectors that reflect the context among words in a sentence. Although these methods efficiently learned context-reflecting vectors, they were unable to capture context-dependent meanings for polysemous words. To address this, Peters et al. \cite{peters2018elmo} introduced ELMo, utilizing Bi-LSTM to generate dynamic word embeddings according to the given context, proving effective in various NLP tasks. Radford et al. \cite{radford2018gpt} proposed GPT, based on Transformer decoder, which learned contextual information during sentence generation, showing outstanding results in text generation. Devlin et al. \cite{devlin2019bert} introduced BERT, based on Transformer encoder, considering bidirectional context, and achieved remarkable performance in a range of NLP tasks including question answering, named entity recognition, etc. Building on these impressive performances, our research aims to utilize BERT, which considers bidirectional contextual information, to convey information to NMT.

\subsection{Integrating PLM into NMT}
Numerous studies have been conducted on integrating PLM into NMT through various approaches.
Ramachandran et al. \cite{Ramachandran2016} proposed initializing parts of NMT model with PLM and subsequently fine-tuning it.
Ding et al. \cite{Ding2018} suggested leveraging PLM embeddings in NMT. The pre-trained embeddings were kept static while being combined with additional embeddings, and only these new embeddings were trained, highlighting the significance of these supplementary embeddings.
Yang et al. \cite{Yang2019} introduced Asymptotic Distillation for transferring knowledge from PLM to NMT. A dynamic switch is employed to utilize the information from PLM in NMT dynamically. Additionally, the importance of differentiating learning rates between PLM and NMT during fine-tuning is conveyed through a proposed rate-scheduled learning.
Zhu et al. \cite{zhu2020incorporating} revealed in their preliminary exploration that using PLM output as an input to NMT proved to be more effective than initializing NMT with PLM parameters followed by fine-tuning. They also proposed BERT-fuse approach, which integrates an additional attention layer that interacts with PLM output in both the encoder and decoder.
Weng et al. \cite{weng2020acquiring} incorporated a Dynamic Fusion Mechanism that considers information from all PLM layers in NMT Encoder. They also proposed a knowledge distillation paradigm for decoder, emphasizing the importance of utilizing multiple layers from PLM.
Xu et al. \cite{xu2021bibert} presented a methodology combining stochastic layer selection with bidirectional pre-training to effectively utilize multi-layers of PLM, underscoring the significance of bidirectional pre-training.
Weng et al. \cite{Weng2022} replaced NMT encoder with PLM and proposed a Layer-wise Coordination Structure to adjust the learning between PLM and NMT decoders. Subsequently, they introduced a segmented multi-task learning method for fine-tuning the pre-trained parameters, highlighting the need to reduce incompatibility between PLM and NMT.

\section{Background}
\subsection{Neural Machine Translation}
The core principle of Neural Machine Translation involves learning the process of converting a given parallel sentence pair \{x, y\} from the source sequence x to the target sequence y. This transformation is facilitated through the Transformer model \cite{vaswani2017attention}. The overall structure of Transformer model is as follows:

\textbf{Input Encoding}: The input sequences x, y first pass through an embedding layer, transforming them into continuous vector representations. Subsequently, position encoding, containing location information, is added to form the final input representation.

\begin{equation}
E_x = \text{Emb}(x) + \text{Pos}(x) 
\end{equation}
\begin{equation}
E_y = \text{Emb}(y) + \text{Pos}(y)
\end{equation}

\textbf{Encoder}: Multiple encoder layers process $E_x$. The i-th encoder layer $H_E^i$ comprises layer normalization (LN) \cite{ba2016layer}, multi-head attention mechanism (MHA), and a feed-forward network (FFN) \cite{vaswani2017attention}. Each layer uses the output of the previous layer as input. $S_E^i$ represents self-attention result of encoder layer.

\begin{equation}
S_E^i = \text{LN}(H_E^{i-1} + \text{MHA}(H_E^{i-1}, H_E^{i-1}, H_E^{i-1}))
\end{equation}
\begin{equation}
H_E^i = \text{LN}(S_E^i + \text{FFN}(S_E^i))
\end{equation}

Here, $H_E^0 = E_x$.

\textbf{Decoder}: Multiple decoder layers process $E_y$. The i-th decoder layer $H_D^i$ consists of LN, MHA, and FFN networks, and also includes attention relationship with the final output of encoder $H_E^N$. $S_D^i$ indicates self-attention result of decoder layer, which is used to calculate attention with the final output of encoder $H_E^N$, obtaining $C^i$.

\begin{equation}
S_D^i = \text{LN}(H_D^{i-1} + \text{MHA}(H_D^{i-1}, H_D^{i-1}, H_D^{i-1}))
\end{equation}
\begin{equation}
C^i = \text{LN}(S_D^i + \text{MHA}(S_D^i, H_E^N, H_E^N)) \\
\end{equation}
\begin{equation}
H_D^i = \text{LN}(C^i + \text{FFN}(C^i))
\end{equation}

Here, $H_D^0 = E_y$.

\textbf{Output}: The final output of decoder is transformed into the probability distribution of the next token through a linear layer and softmax function.

\begin{equation}
\text{Output} = \text{Softmax}(\text{Linear}(H_D^N))
\end{equation}

Here, $H_D^N$ represents the output of the last layer of decoder.

\textbf{Loss Function}: The training objective of NMT is to minimize the difference between the actual target and the model's prediction. The model's Output represents the probability distribution for each token of the target sequence. If the one-hot encoding of the actual target token is $y_{true}$, then the cross-entropy loss is defined as follows:

\begin{equation}
L_{CE} = -\sum y_{\text{true}} \log(\text{Output})
\end{equation}

This loss function measures how close the model's predictions are to the actual target and updates the model's parameters to minimize this loss during training.

\subsection{Pretrained Langue Model}
In recent years, a variety of Pre-trained Language Models (PLMs) such as ELMo \cite{peters2018elmo}, GPT \cite{radford2018gpt}, BERT \cite{devlin2019bert}, XLNet \cite{yang2019xlnet}, BART \cite{lewis2019bart}, and T5 \cite{raffel2020texttotext}, capable of leveraging large-scale monolingual data, have been proposed. There are mainly two methods for training PLMs. 
The first is the auto-regressive approach \cite{radford2018gpt}, where the model operates by predicting the k-th token $z_k$ based on the given context $z_{<k}$ (i.e., the sequence before the k-th token). This can be represented mathematically as $PLM(z_k | z_{<k}; \theta)$.
The second method is the masked language modeling approach introduced by BERT \cite{devlin2019bert}. In this method, random tokens are masked, and these masked tokens are predicted using the surrounding context information. Mathematically, this can be expressed as $PLM(z_m | z_{-m}; \theta)$.

\section{Approach}
\begin{figure*}[tb]
  \centering
  \includegraphics[width=0.8\textwidth]{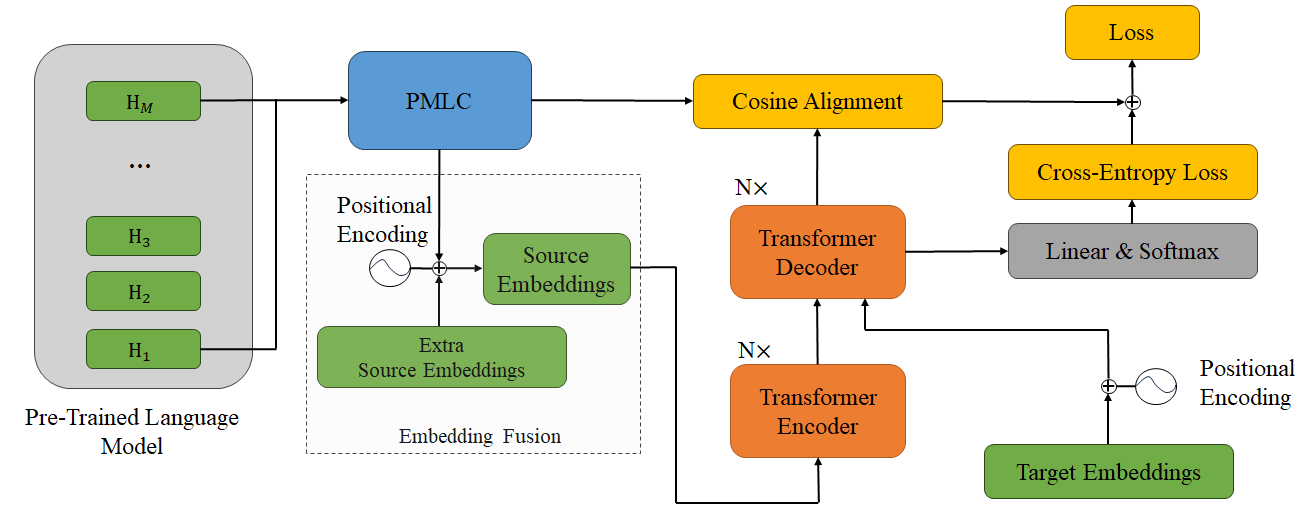}
  \caption{The architecture of the PiNMT model}
  \label{fig:1.png}
\end{figure*}

In this study, we introduce a novel model called PLM-integrated NMT (PiNMT), which integrates PLM into NMT. PiNMT is composed of three components: PLM Multi Layer Converter, Embedding Fusion, and Cosine Alignment. Fig. \ref{fig:1.png} provides a detailed representation of PiNMT architecture, and our approach is designed to address the issues previously raised. Additionally, we describe two training strategies necessary for overcoming these issues: Separate Learning Rates and Dual Step Training.

\subsection{PLM Multi Layer Converter (PMLC)}
PLM are composed of multiple layers, and each layer captures a variety of contextual information \cite{peters2018elmo, jawahar2019does}. Previous research lacked a deep exploration of utilizing the multi-layered nature of PLM \cite{edunov2019pretrained, xu2021bibert}. We introduce a Converter technique to transform the multi-layer of PLM into a suitable source embeddings for NMT model. Additionally, we introduce a Dimensional Compression method to apply the high-dimensional information of PLM output to NMT model with parameter constraints, allowing NMT model to leverage PLM's information more effectively.

\noindent \textbf{Converter}\\
Converter transforms the multi-layer output of PLM into a source embeddings suitable for NMT model.

\subsubsection{Vanilla} 
\begin{figure}[!tb]
    \centering
    \begin{subfigure}{.25\textwidth}
        \centering
        \includegraphics[width=.15\linewidth]{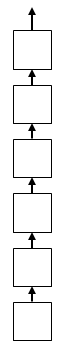}
        \caption{Vanilla}
        \label{fig:a}
    \end{subfigure}%
    \begin{subfigure}{.25\textwidth}
        \centering
        \includegraphics[width=.5\linewidth]{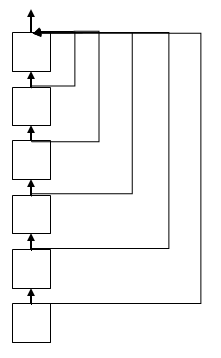}
        \caption{Residual}
    \end{subfigure}
    
    \vspace{1em} 
    
    \begin{subfigure}{.25\textwidth}
        \centering
        \includegraphics[width=.5\linewidth]{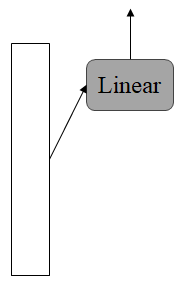}
        \caption{Concat Linear}
    \end{subfigure}%
    \begin{subfigure}{.25\textwidth}
        \centering
        \includegraphics[width=.5\linewidth]{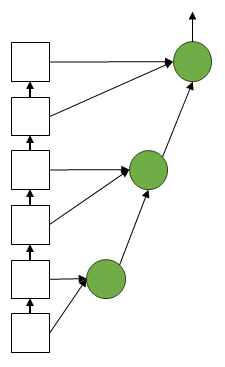}
        \caption{Hierarchical}
    \end{subfigure}
    \caption{Converter}
\end{figure}

As seen in Fig.~\ref{fig:a}, only the final layer of PLM is used for output. This method is based on the theory that the information extracted from the model's final layer is the richest and has the highest contextual understanding \cite{Sun2019}. The last layer of PLM typically captures complex characteristics of the input data and has the capacity to understand sophisticated linguistic features and nuances.

\subsubsection{Residual} 
Inspired by existing research \cite{He2015}, we introduce the concept of shortcut connections to PLM, utilizing their multi layer architecture. The central idea is to combine the outputs of all layers before the last layer with the output of the final layer. This method simply adds the values of existing layers without additional parameters or complex operations, thus being computationally efficient. The equation is as follows, where $H_{PLM}^i$ represents the output of the i-th layer of PLM.

\begin{equation}
\hat{H} = \sum_{i=1}^{M} H_{PLM}^i
\end{equation}

A key difference between our study and ResNet \cite{He2015} lies in the implementation and scope of shortcut connections. ResNet \cite{He2015} modified the foundational model by introducing shortcut connections to its intermediate layers. In contrast, our proposed method maximizes the use of information from all layers without affecting the intermediate structure of the model. This approach minimizes the risk of pre-trained information degradation while allowing for more comprehensive utilization of information.

\subsubsection{Concat Linear} 
We propose a novel approach to resolve incompatibility issues when integrating PLM with NMT model. Our method involves transforming the multi layers of PLM into the input for an NMT model using learnable parameters for additional training. This process effectively converts the multi layers of PLM, enhancing compatibility between the two models. Consequently, it efficiently integrates the robust language understanding capabilities of PLM into NMT model. Specifically, we concatenate the outputs of each layer of PLM and then use a Linear Layer to reduce the dimensions and produce the final output.
\begin{equation}
H' = [H_{PLM}^1; H_{PLM}^2; \dots; H_{PLM}^M]
\end{equation}
\begin{equation}
\hat{H} = W H' + b
\end{equation}
This method is similar to the existing Linear Combination approach \cite{dou2018exploiting} but with several notable differences. The traditional approach combines each layer's output after passing through a linear layer without a bias term. In contrast, our method first concatenates each layer's output and then passes it through a linear layer that includes a bias term. Our proposal introduces a low-dimensional bias parameter to the model. The introduction of this bias parameter enhances the convergence speed during the learning process and significantly contributes to the overall performance improvement of the model.

\subsubsection{Hierarchical} 
As observed in the study by Vaswani et al. \cite{vaswani2017attention}, structuring layers in a deep and complex manner can capture information more effectively. From this perspective, instead of using a simple linear layer, we design a deeper, more intricate Converter. Our objective is to propose a structure that merges nodes hierarchically and deeply, an idea inspired by the research on Hierarchical Aggregation \cite{dou2018exploiting}.

The core concept introduces an aggregation (AGG) node $\hat{H}^i$. This node combines information from either two or three layers using AGG function, depending on specific conditions. AGG function concatenates multiple input layers and forwards them to a Feed Forward Network (FFN). FFN's outcome connects back to the original inputs through a shortcut connection, and the final result is normalized through Layer Normalization (LN).

\begin{equation}
\hat{H}^i = \begin{cases}
\text{AGG}(H^{2i-1}, H^{2i}) & \text{if } i = 1 \\
\text{AGG}(H^{2i-1}, H^{2i}, \hat{H}^{i-1}) & \text{if } i > 1
\end{cases}
\end{equation}

\begin{equation}
\text{AGG}(x, y, z) = \text{LN}(\text{FFN}([x;y;z]) + x + y + z)
\end{equation}

While earlier studies \cite{dou2018exploiting} employed a structure that fed aggregation nodes back into the original backbone. In our approach, we modify this to prevent aggregation node from being re-supplied to the backbone. This adjustment stems from the risk of compromising the pre-trained information in PLM.

\noindent \textbf{Dimensional Compression}\\
The output from Converter encompasses deep, high-dimensional information acquired from large datasets. This high-dimensional data, although rich in meaning, is problematic due to its excessive dimensionality, especially when incorporated into an NMT model with parameter constraints.

To address this, we suggest compressing the output dimensions of the Converter via a Linear layer. This enables NMT model to effectively leverage the information extracted from PLM. Mathematically, this can be represented as:
\begin{equation}
H_{PLM}' = W H_{PLM} + b
\end{equation}

Such a compressed output is then feed into NMT model, allowing for the efficient use of high-dimensional information while optimizing parameter efficiency.

\subsection{Embedding Fusion}
Embedding Fusion is an approach designed to overcome the limitations of fine-tuning PLM. Most PLMs are expansive, making direct fine-tuning a challenging task. Prior studies have frozen the parameters of PLM and integrated it with an NMT model. An Extra Source Embeddings was added to this combined model, which was then fine-tuned to enhance performance \cite{Ding2018}. This method essentially emulates the effects of directly fine-tuning PLM. It alleviates the incompatibility issues between PLM and NMT model, while preserving the pre-trained information. However, research on the effective utilization of Extra Source Embeddings remains scant. Hence, this study proposes an optimized method to harness the potential of Extra Source Embeddings.
\subsubsection{Addition}
To maximize the combination of PLM output and Extra Source Embeddings, we apply a simple yet effective element-wise sum technique. Specifically, PLM output \(H_{PLM}\) and Extra Source Embeddings \(E_x\) are summed to generate a new embedding \(E_x'\). This method preserves features from both sources and effectively combines their information. Formally, this can be represented as:
\begin{equation}
E_x' = H_{PLM} + E_x
\end{equation}

\subsubsection{Multiplication}
We employ an element-wise multiplication technique to more vividly model interactions between the two embeddings. This emphasizes the interdependency and relevance of each feature, resulting in an embedding that closely intertwines the characteristics of the two original embeddings. Mathematically, it is depicted as:

\begin{equation}
E_x' =  H_{PLM} \odot E_x
\end{equation}

\subsubsection{Weighted Sum}
Mere combination might not adequately reflect the relative importance between two embeddings. To address this, we introduce a learnable weight to balance between the two embeddings. Specifically, the weight \(\gamma\) dynamically adjusts the importance of the two embeddings, striving for an optimal combination. This is mathematically captured as:
\begin{equation}
E_x' = \gamma H_{PLM} + (1 - \gamma) E_x
\end{equation}

\subsubsection{Projection}
In the projection approach, each embedding undergoes a linear layer transformation before combining. This ensures both embeddings map onto the same feature space, facilitating efficient information amalgamation and adjustment. This can be mathematically represented as:
\begin{equation}
E_x' = (W_1 H_{PLM} + b_1) + (W_2 E_x + b_2)
\end{equation}

\subsubsection{Concatenation}
The embeddings are concatenated. By directly merging features obtained from various sources through concatenation, the model can utilize the information from both embeddings. However, as the combined embedding might differ in dimensionality from the original space, a Linear Layer is employed for adjustments. The formula for this is:
\begin{equation}
E_x' = Linear([H_{PLM};E_x])
\end{equation}

\subsubsection{Dynamic Switch}
Based on Dynamic switch \cite{Yang2019}, this method features the introduction of a context gate, crucial for the optimal integration of the two embeddings. The context gate, rooted in sigmoid neural network layer, determines the importance of each element within the input vectors received from PLM and Extra Source Embeddings. It's given by the equation:
\begin{equation}
g = \sigma(W H_{\text{PLM}} + U E_x + b)
\end{equation}
Using the computed value of \(g\), the two embeddings are dynamically combined. The value \(g\) adjusts the significance of each embedding, ensuring balanced information integration. This is performed through the equation:
\begin{equation}
h = g \odot h_{\text{PLM}} + (1 - g) \odot E_x
\end{equation}
While the previous research \cite{Yang2019} applied Dynamic Switch to each individual encoder layer, this study focuses solely on Extra Source Embeddings.

\subsection{Cosine Alignment}
In previous studies, methods have been proposed for the effective transfer of knowledge from large models to smaller ones through Distillation \cite{Hinton2015, Yang2019, weng2020acquiring}. Among these, the methodology presented by Yang et al. \cite{Yang2019} for distilling information from PLM to an NMT minimizes the mean-squared-error loss between PLM output and the outputs of NMT encoder or decoder, thereby transferring PLM's knowledge. However, subsequent experimental results indicate that distilling from PLM to an NMT model in low-resource data scenarios either results in suboptimal performance enhancement or even degradation. A primary reason for this phenomenon appears to be the incompatibility between PLM and NMT model.

To address this, Cosine Alignment is proposed. This approach adds cosine similarity between the output of PLM and the last layer of NMT model's decoder to the existing loss function. Since the outputs of PLM and NMT Decoder do not match in sequence length, the average value of each sequence is used.

\begin{gather}
H_{PLM} = \{ h_{PLM}^1, \dots, h_{PLM}^I \} \\
H_D = \{ h_{D}^1, \dots, h_{D}^L \} \\
h_{PLM_{avg}} = \frac{1}{I} \sum_{i=1}^{I} h_{PLM}^i \\
h_{D_{avg}} = \frac{1}{L} \sum_{l=1}^{L} h_{D}^l \\
\begin{aligned}
L_{similarity} &= \text{cosine similarity}(h_{PLM_{avg}}, h_{D_{avg}}) \\
&= \frac{h_{PLM_{avg}} \cdot h_{D_{avg}}}{||h_{PLM_{avg}}|| \times ||h_{D_{avg}}||}
\end{aligned}
\end{gather}

Here, \( H_{PLM} \) represents the output of PLM with a sequence length of \( I \), and \( H_D \) represents the last layer of decoder's output with a sequence length of \( L \). The proposed \( L_{similarity} \) is combined with the traditional cross-entropy loss \( L_{CE} \) to define the final loss function:

\begin{equation}
L = L_{CE} + \alpha L_{similarity}
\end{equation}

In this context, \( \alpha \) serves as a hyper-parameter that adjusts the weights between the two losses.

\subsection{Separate Learning Rates}
PLMs are often large and intricate, and during fine-tuning, there's a risk of losing pre-trained information. Conversely, not fine-tuning PLM can cause incompatibility issues between PLM and NMT model. To mitigate these challenges, research has been proposed to set varying learning rates for different layers \cite{Howard2018, Sun2019, Yang2019}. Previous studies have demonstrated the efficacy of this approach.

In this study, we incorporate the training strategy suggested by Yang et al. \cite{Yang2019} to implement Separate Learning Rates in our PiNMT model. We adjust the learning rate for PLM to be relatively lower compared to that of NMT model. Mathematically, this can be represented as follows:

\begin{equation}
\eta^{PLM} = \rho \times \eta^{NMT}
\end{equation}

Where \( \eta^{PLM} \) denotes the learning rate of PLM, \( \eta^{NMT} \) denotes the learning rate of NMT model, and \( \rho \) indicates the relative coefficient between these learning rates.

\subsection{Dual Step Training}
Many NMT models are trained solely on unidirectional data, which makes it challenging to harness bidirectional linguistic features effectively. However, recent studies \cite{ding2021bidirectional, xu2021bibert} have reported that implementing bidirectional training can substantially enhance NMT performance.

In this study, we refer to the pre-existing training method \cite{xu2021bibert} and apply Dual Step Training to PiNMT model. The core idea behind this approach is to invert the direction of unidirectional data, thereby augmenting it to create bidirectional data (e.g., from En \( \to \) De, we derive En + De \( \to \) De + En).

Utilizing this newly formulated bidirectional data, we conduct a pre-training phase for NMT model. This pre-training facilitates the model in learning bidirectional linguistic attributes, thereby enhancing its generalization capabilities. Subsequently, we fine-tune the pre-trained NMT model with the original unidirectional data to optimize the model's performance for specific translation directions.

\section{Dataset and Baseline Settings}  
\subsection{Dataset}
To validate the efficacy of our proposed methodology, we evaluate it using the IWSLT’14 dataset \cite{cettolo2014iwslt} for the English$\leftrightarrow$German (En$\leftrightarrow$De) language pair. The IWSLT'14 English$\leftrightarrow$German dataset comprises a total of 160K parallel bilingual language pairs, allowing for a quantitative grasp of the model's performance. The distribution ratios of the training, validation, and testing data are detailed in Table~\ref{tab:dataset}. 

\begin{table}[b]
\centering
\caption{Data Distribution}
\label{tab:dataset}
\begin{tabular}{|c|c|}
\hline
\textbf{IWSLT'14 (En$\leftrightarrow$De)} & \textbf{Count} \\ \hline
train & 160239 \\ 
valid & 7283 \\ 
test & 6750 \\ \hline
\end{tabular}
\end{table}

\subsection{Evaluation}
For evaluation metrics, we adopt the commonly used tokenized BLEU Score \cite{papineni2002bleu}. Without the use of Dual Step Training, we set the beam search width to 4 and the length penalty to 0.6. When employing Dual Step Training, the beam search width is increased to 5, and the length penalty is set at 1.0.
\subsection{Settings}
\subsubsection{PLM}
In our study, we choose BiBERT \cite{xu2021bibert} as our PLM. The original BERT model \cite{devlin2019bert} is pre-trained for a single language. However, BiBERT is concurrently trained on both English and German. Built upon the RoBERTa architecture \cite{Liu2019}, BiBERT model consists of 12 layers, has a model dimension of 768, and includes 12 attention heads. The training data for BiBERT combined and shuffled 145GB of German text and 146GB of English text from OSCAR \cite{suarez2019monolingual}. For the text tokenization process, 67GB of randomly sampled English and German texts from the training dataset were used. Using WordPiece tokenizer \cite{wu2016google}, a total of 52K vocabulary was constructed.

\subsubsection{NMT}
For NMT model implementation, we utilize fairseq framework \cite{ott2019fairseq}. As the base model, we choose Transformer model \cite{vaswani2017attention} with transformer\_iwslt\_de\_en settings. This model comprises 6 encoder-decoder layers, has a model dimension of 512, and includes 4 attention heads. Without employing Dimensional Compression, we set the model's dimension to match PLM output, which is 768. However, when applying Dimensional Compression, it is set to 512. Various parameters is used during the training phase to optimize the model's performance. We apply a label smoothing rate of 0.1 to the cross-entropy loss. The maximum tokens per batch are set at 2048, with an update frequency of 16. For learning rate scheduling, we opt for the inverse\_sqrt method. The Beta values for the Adam optimizer are set at (0.9, 0.98), and the initial learning rate is established at 4e-4.

The vocabulary construction for NMT follow BiBERT \cite{xu2021bibert} implementation method. The encoder use a vocabulary size of 52k, matching PLM. The decoder's vocabulary is built based on the IWSLT' 14 data. Without using Dual Step Training, a vocabulary size of 8K is created from the target language data. Conversely, when leveraging Dual Step Training, we construct a 12k-sized English-German joint vocabulary.

\subsubsection{PiNMT}
When applying Dimension Compression, both Concat Linear and Linear Combination methods are compressed using the existing linear layer as they already contain a linear layer. In experiments not using PMLC, PLM is utilized with Vanilla as the base model. For conveying information in Cosine Alignment to encoder, we do not use length averages and instead use the original distillation method. Conversely, when conveying information to decoder in distillation, we use length averages, similar to Cosine Alignment. The hyperparameter $\alpha$ is set to 500.

\section{Results and Analysis}
In this section, we review proposed PiNMT model and two training strategies using the IWSLT'14 En$\leftrightarrow$De dataset.
\subsection{How does Dimensional Compression affect performance?}
Significant findings can be observed in Table~\ref{tab:1}. Compared to using Transformer that solely trains on NMT model with dimensional compression, there is a larger performance enhancement when performing dimensional compression in Vanilla, which utilizes PLM as an input to NMT model. This emphasizes the importance of dimensional compression in the interaction between PLM's information and NMT model. Furthermore, since PLM possesses rich contextual information and high-dimensional features, it suggests that appropriate compression in the model's dimension is required to effectively convey this information to NMT model, which has a limited number of parameters.

\begin{table}[!tb]
    \centering
    \caption{Dimensional Compression Effects on De→En}
    \label{tab:1}
    \begin{tabular}{|l|c|}
        \hline
        \textbf{Models} & \textbf{BLEU} \\
        \hline
        Transformer ($d_{model}$=768) & 33.99 \\
        Transformer ($d_{model}$=512) & 34.12 \\
        Vanilla ($d_{model}$=768) & 37.64 \\
        Vanilla ($d_{model}$=512) & 38.16 \\
        \hline
    \end{tabular}
\end{table}

\subsection{Performance Comparison of PMLC}
We aim to compare the performance of various PMLC methods with Vanilla serving as the baseline model. Firstly, we introduce strategies proposed in previous studies. Linear Combination \cite{dou2018exploiting} linearly combines the outputs of all layers without any bias term. Next, ELMo \cite{edunov2019pretrained} combines the outputs of each layer with learnable scalar weights to generate a new embedding. Additionally, Stochastic Layer Selection \cite{xu2021bibert} involves randomly selecting and utilizing various layers of PLM during the training process.

According to the results presenting in Table~\ref{tab:2}, all models that harness the multi-layer capabilities of PLM outperform basic Vanilla approach. Notably, models equipped with learnable parameters display more significant improvements than their counterparts. This enhancement can be attributed to the learnable parameters' effectiveness in addressing the model's incompatibility issues. By using vector-based methods, which deploy more parameters than scalar approaches, the model achieves exceptional performance. Likewise, Hierarchical approach, with its more profound layer structure, facilitates intricate learning, marking the highest performance among all the discussed strategies. In our subsequent experiments, we delve deeper into analyzing the PMLC, helping us ascertain the most effective strategy.
\begin{table}[!tb]
    \centering
    \caption{Performance Comparison of PMLC on De→En}
    \label{tab:2}
    \begin{tabular}{|l|c|l|c|}
        \hline
        \textbf{Models} & \textbf{BLEU} & \textbf{Existing Models} & \textbf{BLEU} \\
        \hline
        Vanilla ($d_{model}$=512) & 38.16 & Linear Combination & 38.77 \\
        Residual & 38.67 & ELMo & 38.66 \\
        Concat Linear & 38.78 & Stochastic Layer Selection & 38.39 \\
        Hierarchical & 38.96 & & \\
        \hline
    \end{tabular}
\end{table}

\subsection{Performance Comparison of Embedding Fusion}
Embedding Fusion methods are evaluated using Vanilla as the base. Upon examining the results in Table~\ref{tab:3}, we observe performance enhancements in all methods, with the exception of Multiplication approach, compared to Vanilla. Multiplication emphasizes the interaction between the two embeddings. However, its relatively lower performance suggests that the Extra Source Embeddings provides novel information specialized for NMT, which has a comparatively lower correlation with PLM.

The performance improvement noted in all techniques, excluding Multiplication, indicates that the additional learning of Extra Source Embeddings can serve as a solution to incompatibility while preserving the pre-trained information of PLM.

The methods show higher performance in the order of Addition, Weighted Sum, and others, which had fewer parameters. This is due to the insufficient quantity of data required for training parameters that model the interaction between the two embeddings. Consequently, it is evident that choosing an effective model structure based on the amount of data is crucial. In subsequent experiments, Addition method is employed as Embedding Fusion.
\begin{table}[!tb]
    \centering
    \caption{Performance Comparison of Embedding Fusion on De→En}
    \label{tab:3}
    \begin{tabular}{|l|c|l|c|}
        \hline
        \textbf{Models} & \textbf{BLEU} & \textbf{Models} & \textbf{BLEU} \\
        \hline
        Vanilla ($d_{model}$=768) & 37.64 & Projection & 37.97 \\
        Addition & 38.48 & Concatenation & 38.00 \\
        Multiplication & 35.70 & Dynamic switch & 37.99 \\
        Weighted Sum & 38.07 & & \\
        \hline
    \end{tabular}
\end{table}

\subsection{Distillation vs. Cosine Alignment}
In Table~\ref{tab:4}, the results on the left indicate that, generally, both Distillation and Cosine Alignment methods enhance performance in Transformer. However, applying Distillation to NMT decoder results in a performance decline. Moreover, the performance improvements compared to Vanilla model are not substantial for either method.

The results on the right side of Table~\ref{tab:4} show a performance deterioration in both the encoder and decoder when Distillation technique is applied to Vanilla model. Cosine Alignment, on the other hand, improve performance but only in the decoder. Both methods lead to performance degradation in the encoder, primarily because the encoder, already processing PLM's output, experiences an information collision.

Upon closer examination of the reasons for the performance decrease with Distillation in the decoder for both Transformer and Vanilla models, and why it increases with Cosine Alignment. It reveals that Distillation conveys both magnitude and direction of PLM's output vectors to NMT model. In contrast, Cosine Alignment only conveys the vector's directional information. This characteristic alleviates compatibility issues between PLM and NMT model, allowing only the necessary information to be transmitted efficiently.

\begin{table}[!tb]
\renewcommand{\arraystretch}{1.3} 
\centering
\caption{Distillation vs Cosine Alignment on De→En}
\label{tab:4}
\begin{tabular}{|c|c|c|c|}
\hline
\multicolumn{2}{|c|}{\textbf{Transformer ($d_{model}=768$)}} & \multicolumn{2}{|c|}{\textbf{Vanilla ($d_{model}=768$)}} \\
\hline
\textbf{Models} & \textbf{BLEU} & \textbf{Models} & \textbf{BLEU} \\
\hline
Baseline & 33.99 & Baseline & 37.64 \\
Distillation Enc & 35.36 & Distillation Enc & 37.21 \\
Distillation Dec & 33.29 & Distillation Dec & 36.39 \\
Cosine Alignment Enc & 35.33 & Cosine Alignment Enc & 36.95 \\
Cosine Alignment Dec & 35.13 & Cosine Alignment Dec & 38.26 \\
\hline
\end{tabular}
\end{table}

\subsection{Is Separate Learning Rates Strategy Effective?}
We conduct experiments using Vanilla model with a dimension of 712 as the base. Upon reviewing the results in Table~\ref{tab:5}, it is evident that the magnitude of the $\rho$ value significantly impacts performance. If the $\rho$ value is too large, there's a risk of damaging the contextual information from PLM, potentially leading to a decline in performance. Conversely, if the $\rho$ value is too small, incompatibility issues may arise between PLM and NMT model, resulting in potential performance degradation. Therefore, setting an appropriate $\rho$ value is of paramount importance.

Rate-scheduled learning method \cite{Yang2019} demonstrated exemplary performance on extensive resources in past research but fails to replicate the same effect on the low-resource IWSLT'14. This suggests that different datasets, with their unique characteristics and sizes, may require distinct learning rate strategies.

In the experiments, a $\rho$ value of 0.01 yields the most optimal results. Thus, this value will be employed in subsequent experiments.

\begin{table}[!tb]
    \centering
    \caption{Evaluation of Learning Rate Strategies on De→En}
    \label{tab:5}
    \begin{tabular}{|l|r|l|r|}
        \hline
        \textbf{Models} & \textbf{BLEU} & \textbf{Models} & \textbf{BLEU} \\
        \hline
        \( \rho = 0 \)               & 37.64 & \( \rho = 0.01 \)                       & 38.65 \\
        \( \rho = 0.0001 \)         & 38.06 & \( \rho = 0.05 \)                       & 35.66 \\
        \( \rho = 0.0005 \)         & 38.46 & \( \rho = 0.1 \)                        & 33.88 \\
        \( \rho = 0.001 \)          & 38.57 & \( \rho = 1 \)                          & 11.81 \\
        \( \rho = 0.005 \)          & 38.59 & Rate-scheduled learning                 & 11.82 \\
        \hline
    \end{tabular}
\end{table}

\subsection{What is an optimal PMLC for Combination?}
We establish our baseline by using Dimensional Compression to set the model's dimension at 512. Performance is analyzed by combining PMLC with Embedding Fusion (EF), Cosine Alignment (CA), and Separate Learning Rates (SLR). Our primary experiments focus on three PMLC approaches: Hierarchical, Linear Combination \cite{dou2018exploiting}, and Concat Linear.

Based on the results in Table~\ref{tab:6}, when applying Embedding Fusion and Cosine Alignment, Hierarchical approach witnesses a decline in performance. This suggests that the intricate structure of Hierarchical method can lead to excessive complexity when integrating additional techniques, making optimization challenging.

On the other hand, both Linear Combination and Concat Linear methods have a relatively straightforward structure. This simplicity allows for more potential improvements in performance when implementing additional techniques. Notably, Concat Linear method consistently exhibit superior performance across various combinations. This indicates the inherent flexibility of Concat Linear approach, effectively integrating diverse forms of data and techniques.

However, Linear Combination without bias term, despite some enhancements, is comparatively limited. This can be attributed to the bias term providing an additional degree of freedom to the model, enabling it to better capture specific data structures or patterns. Without the bias term, the model might not have the extra information it needs to detect subtle patterns, which could limit its performance gains.

In summary, this research demonstrates that comparatively simpler and more flexible models are better adapted for integrating and combining diverse techniques. It emphasizes the importance of balancing complexity and flexibility when considering technique integration. As a result, we have chosen Concat Linear approach for Converter.

\begin{table}[!tb]
\renewcommand{\arraystretch}{1.3}
\caption{Evaluation of PMLC with Various Methods on De→En}
\label{tab:6}
\centering
\begin{tabular}{|l|c|c|c|}
\hline
\textbf{Models} & \textbf{Linear Combination} & \textbf{Concat Linear} & \textbf{Hierarchical} \\
\hline
Baseline & 38.77 & 38.78 & 38.96 \\
+ EF & 38.91 & 39.07 & 38.68 \\
+ CA & 38.87 & 39.06 & 38.22 \\
+ SLR & 38.97 & 39.12 & 39.17 \\
\hline
+ ALL & 38.92 & 39.71 & 37.97 \\
\hline
\end{tabular}
\end{table}

\subsection{Is Dual Step Training Strategy Effective in PiNMT?}
Using a Transformer Model with a dimension of 512 as the base, we analyze the results of a study applying a Dual Step Training to PiNMT combined with Separate Learning Rates. Observing Table~\ref{tab:7}, it is evident that even the sole application of Bidirectional Pre-training enhances the model's performance. This implies that the model benefits from recognizing the bidirectional characteristics of translation. Additionally, performance improvements are observable with Unidirectional Fine-tuning, indicating that deep learning across extensive data alone is insufficient. It emphasizes the need for learning tailored to specific domains.

\begin{table}[!tb]
    \centering
    \caption{PiNMT performance with Dual Step Training}
    \label{tab:7}
    \begin{tabular}{|l|c|c|}
        \hline
        \textbf{Models} & \textbf{En$\rightarrow$De} & \textbf{De$\rightarrow$En} \\
        \hline
        Transformer (vocab size=12K) & 28.19 & 34.13 \\
        \hline
        PiNMT with Separate Learning Rates & 31.48 & 40.03 \\
        Bidirectional Pre-training & 32.09 & 40.12 \\
        Unidirectional Fine-tuning & 32.20 & 40.43 \\
        \hline
    \end{tabular}
\end{table}

\subsection{Compared with Previous Work}
Table~\ref{tab:8} compares our research with various studies based on the IWSLT'14 En$\leftrightarrow$De dataset. BERT-Fuse \cite{zhu2020incorporating} introduced a new attention layer to augment PLM interactions. UniDrop \cite{wu2021unidrop}  consolidated multiple dropout strategies. R-Drop \cite{wu2021r}  employed a regularization technique utilizing dropout, BIBERT \cite{xu2021bibert} harnessed PLM multi layers and engaged in bidirectional pre-training, and Bi-SimCut \cite{gao2022bi} enhanced performance by seamlessly integrating data augmentation and bidirectional pre-training. Compared to Transformer, our method showcases an ascent of 5.16 in the BLEU score, and it exceeds the prior peak performance set by Bi-SimCut by an additional 1.55 BLEU score. These results provide compelling evidence for the effective resolution of the identified challenges.

\begin{table}[!tb]
    \centering
    \caption{Comparison with Existing Models}
    \label{tab:8}
    \begin{tabular}{|l|c|c|c|}
        \hline
        \textbf{Models} & \textbf{En$\rightarrow$De} & \textbf{De$\rightarrow$En} & \textbf{Average} \\
        \hline
        Transformer (vocab size=12K) & 28.19 & 34.13 & 31.16 \\
        BERT-Fuse & 30.45 & 36.11 & 33.28 \\
        UniDrop & 29.33 & 36.41 & 32.87 \\
        R-Drop & 30.72 & 37.25 & 33.99 \\
        BIBERT & 30.45 & 38.61 & 34.53 \\
        Bi-SimCut & 31.16 & 38.37 & 34.77 \\
        \hline
        Our Model & 32.20 & 40.43 & 36.32 \\
        \hline
    \end{tabular}
\end{table}

\section{Conclusion}
We focused on the incompatibility issues arising when integrating PLM into NMT. While PLMs were designed to understand and generate text within a single language, NMT models were tasked with translating between different languages. The inherent differences between these tasks led to incompatibility issues. To address these, we proposed PiNMT model, incorporating key components like PMLC, Embedding Fusion, and Cosine Alignment.

We designed PiNMT model to leverage the rich contextual insights from PLM, all the while overcoming the challenges of their integration with NMT. In addition, to make model training more effective, we incorporated strategies like Separate Learning Rates and Dual Step Training. By adopting these methodologies, we achieved SOTA performance on the IWSLT'14 En$\leftrightarrow$De dataset.

As with any research, our methodology can be further refined. Further tests considering diverse languages and scales, as well as extended research, are necessary. In conclusion, this study served as a foundational step in strengthening the linkage between PLM and NMT, laying a critical groundwork for future advancements in translation models.

\bibliographystyle{IEEEtran}
\bibliography{references}
\end{document}